\documentclass[conference]{IEEEtran}
\IEEEoverridecommandlockouts
\usepackage{graphicx} 
\usepackage{pifont}
\usepackage{amsmath}
\usepackage{amssymb}  
\usepackage{amsfonts}
\usepackage{algorithmic}
\usepackage{array}
\usepackage{makecell}

\usepackage{multicol}
\usepackage{diagbox}
\usepackage{textcomp}
\usepackage{stfloats}
\usepackage{url}
\usepackage{verbatim}
\usepackage{graphicx}
\usepackage{caption}
\usepackage{subcaption}
\usepackage[ruled, vlined, linesnumbered]{algorithm2e}
\usepackage{cite}
\usepackage{booktabs}
\usepackage{amssymb,amsfonts,bm}
\usepackage{amsmath,tikz}
\usepackage{color}
\usepackage{mathtools}
\usepackage[hidelinks]{hyperref} 
\usepackage{rotating}
\usepackage{blkarray}
\usepackage{physics}
\usetikzlibrary{arrows}
\usepackage{makecell}

\usepackage{amsthm}
\usepackage{comment}
\usepackage{multirow}
\usepackage{geometry}
\graphicspath{{images/}}

\begin{document}


\newcommand{\Dweb}{\mathcal{D}_{\text{web}}}
\newcommand{\Xhat}{\hat{\mathbf{X}}}
\newcommand{\yhat}{\hat{y}}
\newcommand{\Vi}{v_i}
\newcommand{\Zi}{\mathcal{Z}_i}
\newcommand{\Det}{\mathrm{Det}}
\newcommand{\EKF}{\mathrm{EKF}}

\title{\LARGE \bf
3D UAV Trajectory Estimation and Classification from Internet Videos via Language Model
}

\author{Haoxiang Lei, Daotong Wang, Shenghai Yuan, Jianbo Su*}


\maketitle

\begin{abstract}

Reliable 3D trajectory estimation of unmanned aerial vehicles (UAVs) is a fundamental requirement for anti-UAV systems, yet the acquisition of large-scale and accurately annotated trajectory data remains prohibitively expensive. In this work, we present a novel framework that derives UAV 3D trajectories and category information directly from Internet-scale UAV videos, without relying on manual annotations. First, language-driven data acquisition is employed to autonomously discover and collect UAV-related videos, while vision-language reasoning progressively filters task-relevant segments. Second, a training-free cross-modal label generation module is introduced to infer 3D trajectory hypotheses and UAV type cues. Third, a physics-informed refinement process is designed to impose temporal smoothness and kinematic consistency on the estimated trajectories. The resulting video clips and trajectory annotations can be readily utilized for downstream anti-UAV tasks. To assess effectiveness and generalization, we conduct zero-shot transfer experiments on a public, well-annotated 3D UAV benchmark. Results reveal a clear data scaling behavior: as the amount of online video data increases, zero-shot transfer performance on the target dataset improves consistently, without any target-domain training. The proposed method closely approaches the current state-of-the-art, highlighting its robustness and applicability to real-world anti-UAV scenarios. Code and datasets will be released upon acceptance.



\end{abstract}


\section{Introduction}


Unmanned aerial vehicles (UAVs) have become increasingly prevalent across civilian and commercial domains, which in turn has intensified concerns regarding airspace safety and counter-UAV defense. A fundamental capability for such systems is the accurate estimation of UAV trajectory in 3D space, as reliable 3D trajectory information is essential for interception planning, and risk-aware decision-making \cite{lei2024audioarraybased3duav,avfdti}. Despite sustained research interest, progress in 3D UAV trajectory estimation remains constrained by the limited availability of large-scale, high-quality annotated 3D UAV data.



Most \textbf{existing} approaches for UAV trajectory estimation and classification rely heavily on large amounts of well-annotated data \cite{lei2024audioarraybased3duav,avdtec,avfdti,ESWA_Lei}, whose collection is expensive, time-consuming, and inherently difficult to scale. Although multiple public datasets support 2D UAV trajectory analysis \cite{antiuav1,antiuav3,wangantidataset}, the absence of depth information significantly constrains their applicability in practical counter-UAV deployments. In contrast, datasets providing full 3D UAV trajectories are exceedingly limited. Among the few available resources, MMAUD \cite{yuan2024MMAUD} offers high-quality and precise 3D annotations. Nevertheless, its construction depends on costly sensors, such as high-precision LiDAR systems, together with extensive manual annotation efforts, which collectively impose substantial obstacles to the large-scale development of 3D UAV trajectory datasets.

\begin{figure}[t]
\centering
\includegraphics[width=3.4in]{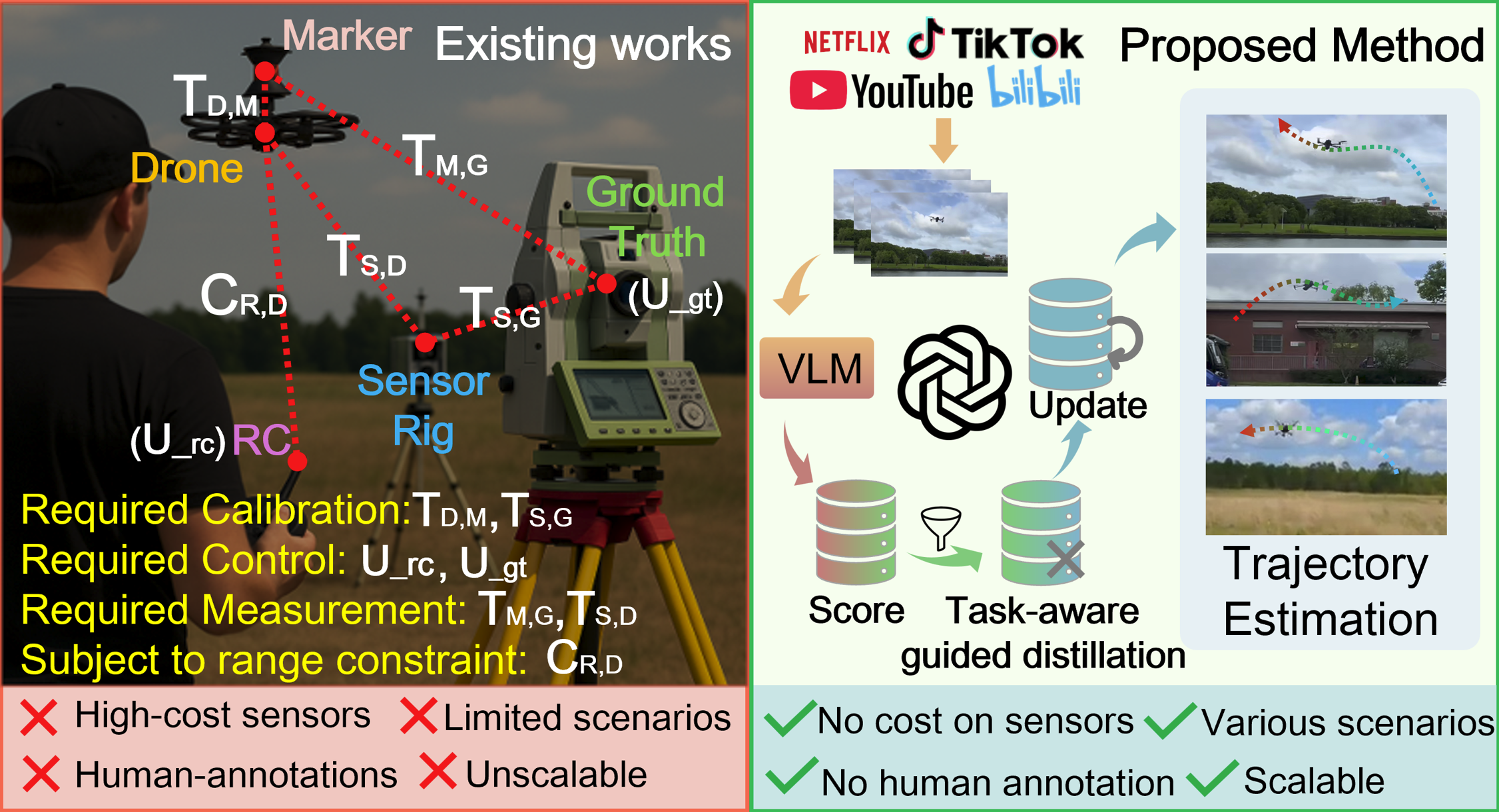}
\caption{The comparison between the acquisition of existing Anti-UAV datasets, and our proposed methods for anti-UAV datasets acquisition. }
\label{motivation}
\end{figure}

The \textbf{challenge} stems from two fundamental factors. First, UAV videos collected from online platforms are highly noisy, frequently including first-person recordings or severe camera motion, which obscures reliable observation of UAV dynamics. Second, acquiring accurate 3D trajectory annotations remains prohibitively expensive. For example, the MMAUD dataset \cite{yuan2024MMAUD} achieves high-precision labeling through Leica MS60 surveying equipment and multi-sensor systems; however, such configurations require costly hardware, meticulous calibration, and substantial manual effort, rendering them inherently unsuitable for large-scale deployment. Together, these constraints highlight the need for a new approach capable of converting  Internet videos into reliable 3D trajectory resources without relying on specialized sensors or human annotation.



To \textbf{address} these challenges, a novel framework that converts raw Internet-scale UAV videos into accurate 3D trajectories and UAV type labels, \emph{without manual annotations or sensing hardware}, is proposed. The framework is guided by three key principles. First, \textbf{language-driven data acquisition}, in which an agentic large language model (LLM) autonomously discovers UAV-related videos and cooperates with vision-language models (VLMs) to progressively remove irrelevant or noisy content. Second, a  \textbf{training-free cross-modal label generation} integrates vision–language reasoning with geometric constraints to infer zero-shot 3D trajectory hypotheses and category information. Third, \textbf{physics-informed refinement} is employed to impose temporal consistency and kinematic feasibility through sequential estimation. The resulting video clips and trajectory annotations can be readily utilized for downstream anti-UAV tasks. We validate the framework via zero-shot transfer on the 3D MMAUD benchmark \cite{yuan2024MMAUD}, which closely approaches the current state-of-the-art. Importantly, the proposed framework demonstrates a clear \textbf{scaling behavior}: as the amount of online video data increases, zero-shot transfer performance on the target dataset improves consistently, without any target-domain training. Our main contributions are summarized as follows:

\begin{enumerate}

\item A \textbf{scalable framework} that derives UAV 3D trajectories and category labels directly from Internet-scale videos \emph{without manual annotation or costly sensors} is proposed, and reveals a clear data scaling behavior.

\item A \textbf{language-driven data acquisition} is designed, in which an agentic LLM works in conjunction with vision--language models to automatically retrieve and refine task-relevant UAV video content.

\item A \textbf{training-free cross-modal label generation and physics-informed refinement} pipeline is introduced that integrates vision--language cues with sequential estimation to ensure temporal coherence and kinematic plausibility.

\item We conduct a comprehensive \textbf{zero-shot evaluation} on the 3D MMAUD benchmark, achieving competitive performance, closely approaching the current state-of-the-art, while establishing a reproducible and scalable paradigm for anti-UAV perception.
\end{enumerate}

\section{Related works}

\subsection{Annotated Datasets for Anti-UAV Research}

Several datasets have been developed for anti-UAV research, including the Anti-UAV benchmarks~\cite{antiuav1,2ndantiuav,antiuav3} and DUT Anti-UAV~\cite{dut-anti}, which provide large-scale annotated videos for UAV tracking. However, these datasets rely on labor-intensive frame-level labeling and are limited to 2D tracking in RGB or thermal imagery, restricting their applicability to real-world 3D trajectory estimation.

To enable 3D annotation, Zheng et al.~\cite{zhaoshiyudataset} proposed a multi-camera UAV dataset using synchronized and calibrated RGB cameras. While this approach supports 3D reconstruction, it incurs high hardware and calibration costs and has not been publicly released, limiting reproducibility. The MMAUD dataset~\cite{yuan2024MMAUD} offers a high-quality 3D benchmark by employing high-precision Leica MS60 ground truth together with LiDAR, stereo cameras, and a microphone array. Nevertheless, its construction depends on expensive multi-sensor systems and meticulous calibration.

In summary, existing anti-UAV datasets either focus on 2D tracking with costly manual annotation or rely on complex sensing infrastructures for 3D labeling, posing major challenges for scalable 3D UAV trajectory estimation.

\begin{figure*}[t]
\centering
\includegraphics[width=17cm]{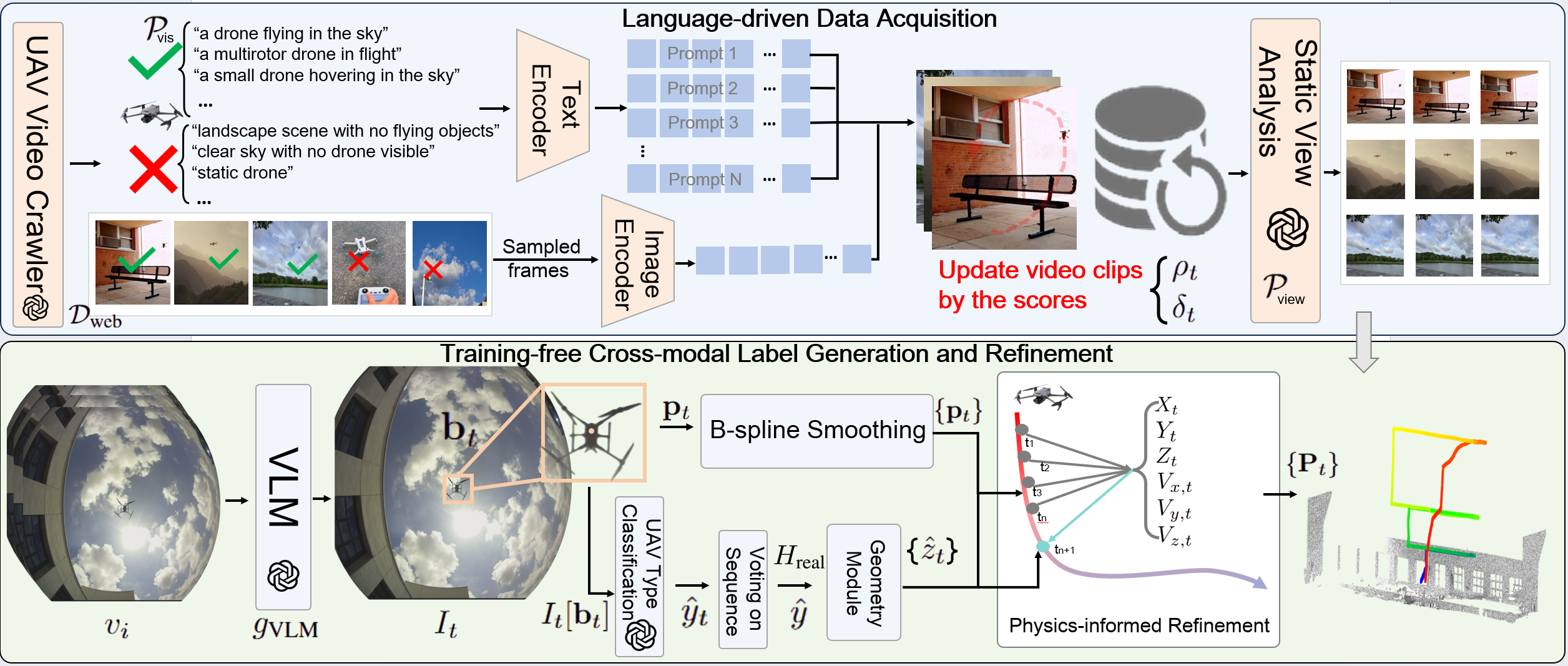}
\caption{The overall framework consists of three main parts, the one on the top is the Language-driven Data Acquisition module for high-quality anti-UAV data. The others on the bottom are training-free cross-modal label generation for 3D UAV trajectory and classification, and the physics-informed refinement. The final output is the 3D trajectory of the UAV, and the LiDAR point cloud is used for visualization only, not for processing.}
\label{fig:network}
\vspace{-1em}
\end{figure*}

\subsection{Dataset-Driven Trajectory Estimation Approaches}


Recent studies on anti-UAV trajectory estimation have largely relied on high-quality, well-annotated datasets or sensor-rich acquisition systems. For example, YOLO-RAW~\cite{YOLO_RAW} improves UAV detection robustness in adverse weather scenarios, while AV-DTEC~\cite{avdtec} and AV-FDTI~\cite{avfdti} exploit LiDAR-generated pseudo labels to train audio–visual or audio-only models for UAV detection and trajectory inference. Similarly, AAUTE~\cite{lei2024audioarraybased3duav} estimates UAV 3D trajectories from audio signals using LiDAR-based supervision and smoothing techniques, achieving strong performance on the MMAUD dataset.

Despite these advances, existing approaches fundamentally depend on curated datasets with precise annotations or sophisticated sensing infrastructures. To the best of our knowledge, none of them address the problem of extracting 3D UAV trajectories directly from unlabeled Internet-scale videos, leaving scalability and real-world deployment largely unexplored.

\section{Proposed Framework}

The proposed framework for extracting UAV trajectories and category labels from web-scale video data consists of three key principles: language-driven data acquisition, training-free cross-modal label generation, and physics-informed refinement, as shown in Fig.~\ref{fig:network}.

\subsection{Language-driven Data Acquisition}


To construct a large-scale data foundation, our framework begins with a language-guided agentic crawler. An autonomous large language model (LLM) retrieves UAV-related videos from multiple open platforms using generic textual queries (e.g., ``UAV flying''), forming an unlabeled corpus
\(
\mathcal{D}_{\text{web}}=\{v_i \mid v_i \in \{\text{YouTube}, \text{TikTok}, \text{Bilibili}, \ldots\}\},
\quad
v_i=\{I_t\}_{t=1}^{T_i},\ I_t\in\mathbb{R}^{H\times W\times 3}.
\)
Here, $T_i$ denotes the number of frames and each $I_t$ is an RGB image.
Although this strategy enables diverse UAV scenarios to be collected at scale, the raw corpus contains substantial noise, including first-person onboard footage and irrelevant instructional content.

To overcome this challenge, the LLM coordinates with vision-language reasoning to perform fine-grained filtering. Specifically, the filtering process is conducted in a progressive manner as shown in upper part of Fig.~\ref{fig:network}. For each video clip, the VLM is first applied to sampled frames to assess UAV visibility by computing frame--prompt relevance scores. Clips with insufficient relevance are discarded at this stage. For the remaining candidates, the same model is further used to evaluate viewpoint characteristics by contrasting static-view and dynamic-view prompts. Frame-level scores are aggregated at the video level to obtain both relevance and viewpoint confidence statistics. The LLM then operates on a structured summary composed of textual metadata together with these aggregated VLM scores, and outputs a discrete decision in a predefined format (accept or reject). This staged design enables fully autonomous filtering while explicitly enforcing the progressive language-driven data priors.


First, UAV visibility is evaluated using a set of visibility-specific language prompts
\(
\mathcal{P}_{\text{vis}}=\{p_j\}_{j=1}^{M},
\)
where each prompt explicitly describes the presence or absence of a visible UAV in the scene.
A vision--language model (VLM) is formulated as a mapping
\(
f_{\text{VLM}}:\mathbb{R}^{H\times W\times 3}\times\mathbb{S}\rightarrow\mathbb{R},
\)
which computes a relevance score between an image frame $I_t$ and a prompt $p_j$ using cosine similarity. For each frame, the maximum score over all visibility prompts is taken as the UAV visibility score, while the margin between the highest and lowest scores is used as a relative confidence measure, reflecting the ambiguity of UAV presence. If the margin is less than threshold \(\tau\), the frame will be discarded.


Second, beyond visibility assessment, the same VLM is used to evaluate viewpoint characteristics.
We define a set of viewpoint-specific prompts
\(
\mathcal{P}_{\text{view}}=\{p_{\text{sta}},p_{\text{dyn}}\},
\)
where $p_{\text{sta}}$ corresponds to static-view recording and $p_{\text{dyn}}$ denotes dynamic-view filming.
Applying
\(
f_{\text{VLM}}(I_t,\mathcal{P}_{\text{view}})
\)
allows the framework to distinguish static-view sequences. Only clips with both a high UAV relevance score and a high static-view confidence are retained. The static-view sequences denote that UAV motion is observable against a stable background, from dynamic-view sequences dominated by camera motion.

Frame-level visibility and viewpoint scores are aggregated at the video level and jointly provided to the LLM, which produces a discrete accept--reject decision. This progressive filtering strategy retains only UAV-relevant and static-view segments for downstream trajectory estimation and classification. Further implementation details are provided in Section~IV.

\subsection{Training-free Cross-modal Label Generation}

The second component of the framework is \emph{training-free cross-modal label generation}, which exploits vision--language reasoning to extract geometric and categorical cues from video frames.

We employ a mixture-of-experts strategy consisting of $K$ heterogeneous detection functions
\[
\{ g_{\text{det}}^{(k)} \}_{k=1}^{K}, \quad
g_{\text{det}}^{(k)} : \mathbb{R}^{H \times W \times 3} \rightarrow \mathbb{R}^{4},
\]
each mapping a frame $I_t$ to a candidate bounding box
\(
\mathbf{b}_t^{(k)} = (x_t^{(k)}, y_t^{(k)}, w_t^{(k)}, h_t^{(k)})^\top.
\) The experts may include open-vocabulary grounding models and lightweight UAV detectors, such as \cite{dino, groundedsam,anti-UAV_dataset,dut-anti}.


For each frame, candidate boxes $\{ \mathbf{b}_t^{(k)} \}$ are clustered 
according to Intersection-over-Union (IoU) using a threshold $\tau'$. 
Formally, two boxes belong to the same cluster if their IoU exceeds $\tau'$. Let $\mathcal{C}_t^{(j)}$ denote the $j$-th IoU cluster at frame $t$, and let $\left|\mathcal{C}_t^{(j)}\right|$ be the number of supporting experts. A cluster is retained only if it is supported by at least two experts as \(
\left|\mathcal{C}_t^{(j)}\right| \ge 2. \) If multiple clusters satisfy this condition, we select the one with the highest aggregated confidence score as \(
j^{\ast} = \arg\max_{j} \; s_t^{(j)}, \)
where the cluster confidence $s_t^{(j)}$ is computed as the average confidence of the supporting expert predictions within the cluster,
\[
s_t^{(j)} = \frac{1}{\left|\mathcal{C}_t^{(j)}\right|}
\sum_{(k)\in \mathcal{I}_t^{(j)}} c_t^{(k)}.
\]
Here, $\mathcal{I}_t^{(j)}$ denotes the index set of experts whose boxes belong to
cluster $\mathcal{C}_t^{(j)}$, and $c_t^{(k)}$ is the confidence score reported by the $k$-th expert.For the selected cluster $\mathcal{C}_t^{(j^{\ast})}$, the fused bounding box is obtained by averaging:
\begin{equation}
\mathbf{b}_t =
\frac{1}{\left|\mathcal{C}_t^{(j^{\ast})}\right|}
\sum_{\mathbf{b} \in \mathcal{C}_t^{(j^{\ast})}} \mathbf{b}.
\end{equation}



In parallel, UAV classification is performed on the cropped region defined by $\mathbf{b}_t$.
At each frame, a vision--language classifier outputs a categorical prediction together with a self-reported confidence score,
\(
\{\hat{y}_t, c_t\} = g_{\text{VLM}}^{\text{cls}}\big(I_t[\mathbf{b}_t]\big),
\quad
\hat{y}_t \in \{1,\dots,C\},\ c_t \in [0,1],
\)
where $C$ is the number of UAV categories.
Frame-level predictions are temporally aggregated to obtain a clip-level label.
Majority voting determines the dominant category $\hat{y}$, while a consistency rule is applied: if the dominant label maintains an agreement ratio above a threshold $\rho$ over $N$ consecutive frames, its confidence is reinforced; otherwise, the result is marked as uncertain.
This strategy combines self-reported confidence with temporal consistency to suppress spurious predictions.

Furthermore, besides category inference, we query the same VLM to estimate the physical size prior from the cropped region, \( H_{\text{real}} = g_{\text{VLM}}^{\text{size}}\big(\hat{y}\big),   H_{\text{real}} \in \mathbb{R}_{+}\).
Given $H_{\text{real}}$ and the bounding-box height $h_t$, a coarse monocular depth cue is estimated as
\(
    \hat{z}_t = ({f_y H_{\text{real}}})/{h_t}, \hat{z}_t \in \mathbb{R},
\)
where $f_y$ denotes the vertical focal length of the camera, which is estimated by the deep-learning-based single-image camera calibration DeepCalib~\cite{single_cali}.

The coarse depth estimate is further refined through a physics-informed refinement process. By combining the 2D trajectory $\{\mathbf{p}_t\}$ with the estimated depths $\{\hat{z}_t\}$, the framework generates zero-shot 3D pseudo-labels
\(
\hat{\mathbf{P}}_t = (u_t, v_t, \hat{z}_t)^\top \in \mathbb{R}^3.
\)
Here, $(u_t, v_t)$ are image-plane coordinates and $\hat{z}_t$ represents a coarse geometric depth, such that $\hat{\mathbf{P}}_t$ forms a mixed pixel--depth representation rather than a full camera-coordinate position. These pseudo-labels serve as weak observations for subsequent refinement, enabling trajectory reconstruction and UAV type inference without human annotation. Further implementation details are provided in Section~IV.

\subsection{Physics-informed Refinement}



The final component of the proposed framework is a physics-informed refinement module, whose goal is to enforce temporal smoothness and kinematic plausibility in the recovered UAV 2D trajectories and coarse depth estimation.
Rather than treating frame-wise geometric estimates independently, we formulate trajectory refinement as a sequential state estimation problem that combines simple motion dynamics with noisy per-frame observations.

The latent state at time $t$ is defined as
\(
\mathbf x_t = [\,X_t,\,Y_t,\,Z_t,\,V_{x,t},\,V_{y,t},\,V_{z,t}\,]^\top \in \mathbb{R}^6,
\)
which jointly represents the 3D position and velocity of the UAV.
The corresponding observation vector is
\(
\mathbf z_t = [\,u_t,\,v_t,\,h_t\,]^\top \in \mathbb{R}^3,
\)
where $(u_t,v_t)$ denote the image-plane center of the detected bounding box and $h_t$ its pixel height.


The relationship between the latent state and image measurements is described by a perspective projection model:
\begin{equation}
\begin{aligned}
\mathbf h(\mathbf x_t) &=
\begin{bmatrix}
c_x + f_x \tfrac{X_t}{Z_t}\\[3pt]
c_y + f_y \tfrac{Y_t}{Z_t}\\[3pt]
H_{\text{real}}\, \tfrac{f_y}{Z_t}
\end{bmatrix}, \\[6pt]
\mathbf H_t &= \left.\frac{\partial \mathbf h}{\partial \mathbf x}\right|_{\mathbf x_t},
\end{aligned}
\label{eq:cross3}
\end{equation}
where $\mathbf h:\mathbb{R}^6 \rightarrow \mathbb{R}^3$ and $\mathbf H_t \in \mathbb{R}^{3\times 6}$ is the associated Jacobian.
Note that the observation function depends only on the position components, while velocity affects the estimation through temporal propagation.


For temporal evolution, we adopt a near-constant velocity assumption over the frame interval $\Delta t$, leading to the prediction step
\begin{equation}
\mathbf x_{t+1|t} =
\begin{bmatrix}
\mathbf I_3 & \Delta t\,\mathbf I_3\\
\mathbf 0   & \mathbf I_3
\end{bmatrix}
\mathbf x_t ,
\label{eq:task01}
\end{equation}
where $\mathbf F\in\mathbb{R}^{6\times6}$ encodes kinematic continuity.
This assumption is sufficient for short inter-frame intervals in video data, where the effect of acceleration on projected image measurements is dominated by measurement noise and geometric uncertainty, and any unmodeled higher-order dynamics are implicitly captured by the process noise.


At each time step, a coarse depth cue \(\hat z_{t+1}\) is obtained from the \(({f_y H_{\text{real}}})/{h_{t+1}}\), which is consistent with the third component of the observation model. The state estimate is then corrected via
\begin{equation}
\mathbf x_{t+1}
=
\mathbf x_{t+1|t}
+
\mathbf K_{t+1}
\bigl(\mathbf z_{t+1} - \mathbf h(\mathbf x_{t+1|t})\bigr),
\label{eq:cross4}
\end{equation}
where the adaptive gain $\mathbf K_{t+1}$ is determined by the process noise $\mathbf Q$ and measurement noise $\mathbf R$.


Through recursive prediction and correction, the physics-informed module fuses generated pseudo-observations $\{\hat{\mathbf P}_t\}$ with motion priors.
The position components of the refined state yield the final 3D trajectory
\(
\mathbf P_t = (X_t, Y_t, Z_t)^\top \in \mathbb{R}^3
\)
in the camera coordinate system.
In this way, noisy frame-level pseudo-trajectories are transformed into temporally consistent and physically feasible trajectories that respect both camera geometry and UAV motion characteristics.

\section{Experiments and Performance}

\begin{figure}[t]
\centering
\includegraphics[width=3.2in]{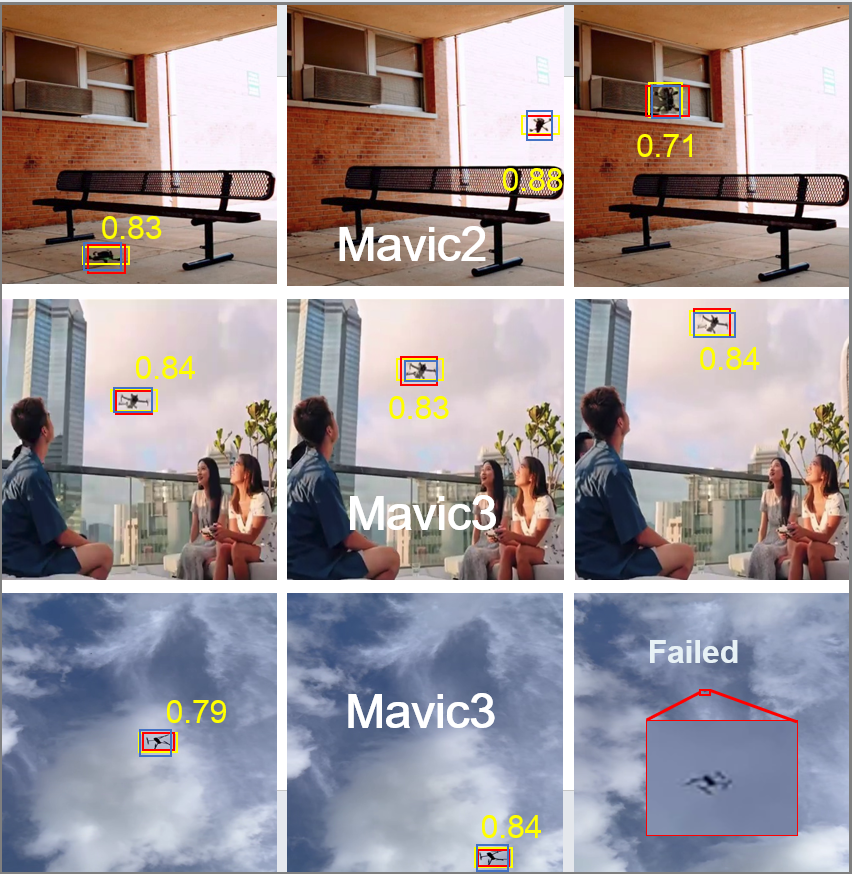}
\caption{The red bounding boxes represent the output of the benchmark solution from Anti-UAV \cite{anti-UAV_dataset}, blue boxes represent the Grounded SAM output \cite{groundedsam}, and the yellow boxes represent the lightweight UAV detector output \cite{drone_yolov8}. }
\label{youtube_result}
\end{figure}

\subsection{Implementation Details}
 
While our framework is general and can integrate diverse LLMs, VLMs, and physics-informed refinement strategies, in practice, we adopt representative off-the-shelf models to demonstrate feasibility and reproducibility.


\textbf{Language-driven Data Acquisition:}
We implement the LLM-based acquisition using GPT-4o (realtime-preview-2025-06-03) in an agentic setting to issue search queries and collect candidate online videos.
For weak semantic guidance and task-aware distillation, we apply CLIP (ViT-B/32)~\cite{radford2021learningtransferablevisualmodels} to video frames sampled at 2~Hz with contrasting prompts (e.g., ``a small drone hovering in the sky'' versus ``clear sky without a visible drone''). Both frames and prompts are embedded by CLIP, and cosine similarity is computed. The maximum similarity score is used as the UAV relevance score, while the margin between the highest and lowest scores serves as a confidence indicator. Frames aligned with ``no drone in view'' are removed first, followed by frames with low confidence (margin $<\tau=0.3$), which filters out ambiguous or irrelevant content. CLIP is further employed with static--dynamic viewpoint prompts to distinguish static-view recordings from sequences dominated by camera motion. Only clips satisfying both UAV relevance ($\geq0.7$) and static-view confidence ($\geq0.3$) are retained, yielding a curated dataset for downstream trajectory estimation and classification. Overall, this process produces approximately 200{,}000 seconds of video across 2{,}245 sequences.


\textbf{Training-free cross-modal label generation:}  
For 2D trajectory estimation, we instantiate the cross-modal label generation using a mixture of three ($K=3$) expert models: Grounding SAM~\cite{groundedsam}, a lightweight pretrained Drone detector~\cite{drone_yolov8}, 
and the benchmark method from the Anti-UAV dataset~\cite{anti-UAV_dataset}, as shown in Fig.~\ref{youtube_result}. For each frame, candidate bounding boxes from the $K=3$ experts 
are clustered according to Intersection-over-Union (IoU) 
with a threshold $\tau' = 0.5$. Clusters supported by at least two experts are retained; we select the one with the highest aggregated confidence score as the final cluster. For the retained cluster, the fused bounding box is obtained by averaging the boxes within that cluster. The barycenters of the fused boxes are then extracted to form discrete 2D trajectory hypotheses. To mitigate detection jitter, the barycenters of selected boxes are smoothed using cubic B-splines with order $\theta=3$ and a uniform knot spacing of 5 frames ($\eta$=5), yielding continuous 2D trajectories.

For UAV classification, cropped bounding-box regions are queried by GPT-4o (realtime-preview-2025-06-03) at 0.1~Hz, and clip-level labels are obtained via majority voting with a window size of $N=5$ and threshold $\rho=0.9$.


\textbf{Physics-Informed Refinement:}
To enforce temporal coherence and suppress depth noise, we implement the physics-informed refinement module using an Extended Kalman Filter (EKF). The EKF integrates per-frame geometric cues with a near-constant velocity motion model, producing smooth and physically plausible 3D trajectories. While EKF serves as a practical instantiation in this work, the framework readily accommodates more advanced sequential estimation techniques.


\subsection{Zero-shot Transfer on the well-annotated dataset}

Based on our training-free method for online UAV video 3D trajectory generalization, we choose the only public 3D anti-UAV dataset--MMAUD \cite{yuan2024MMAUD} to evaluate our 3D Trajectory estimation and classification method. 

Using the cross-modal label generation together with the physics-informed refinement, 
we obtain temporally consistent 3D trajectories in the camera coordinate system,
\(
\mathbf{P}_t = (X_t, Y_t, Z_t)^\top \in \mathbb{R}^3.
\)
To maintain consistency with the web-scale setting, we estimate intrinsics by DeepCalib \cite{single_cali} rather than using the provided calibration for trajectory inference; only the provided RGB--Leica extrinsics are applied to transform the estimated trajectories into the MMAUD world frame, enabling direct comparison with the ground-truth trajectories $\mathbf{P}_t^{\mathrm{GT}}$. 
This transformation is used \textbf{solely} for benchmark-level comparison as shown in Table~\ref{tab:3dcomparison}. The proposed web-scale framework itself does not require camera extrinsics and operates entirely in the camera coordinate system. Trajectory accuracy is measured by the mean squared error over the full sequence:
\(
e_{\text{3D}} =
\frac{1}{T}
\sum_{t=1}^{T}
\left\|
\mathbf{P}_t - \mathbf{P}_t^{\mathrm{GT}}
\right\|_2^2 .
\)

\begin{table*}[t]
\footnotesize
\renewcommand*{\arraystretch}{0.9}
\centering
\caption {The comparison of performance on MMAUD dataset \cite{yuan2024MMAUD} in m.}
\vspace{-5pt}
\renewcommand{\arraystretch}{1.0} 
\resizebox{0.95\textwidth}{!}{
\begin{tabular}{lcccccccccc}
\hline
\toprule
\textbf{Methods}  & \textbf{Year}  & \textbf{Category} &  \textbf{Modal}  & \multicolumn{4}{c}{\textbf{Position Estimation}}  & \textbf{Classification} & FPS  \\
\cmidrule(lr){3-4} \cmidrule(lr){5-8}  
\ &  & (Training)&(Inference) & $D_x$ & $D_y$ & $D_z$ & $e_{\text{3D}}$ & \textbf{Accuracy}  \\
\midrule

\textbf{TAME} \cite{TAME}& 2023 & Supervised & Audio-Only & \textbf{0.11} & 0.30 & 0.34 & 0.55 & \underline{0.960}  & \underline{29.72}\\

\textbf{AV-PED} \cite{av-ped}& 2023 & Self-Supervised & Audio+Image & 0.31 & 0.43 & 0.52 & 0.87 & 0.896 & 12.54\\

\textbf{AV-FDTI} \cite{avfdti}& 2024 & Supervised & Audio+Image & \underline{0.13} & 0.23 & 0.36 & 0.53 & 0.920 & 11.57 \\

\textbf{YOLOv10} \cite{yolov10} & 2024 & Supervised & Image-Only & 0.23 & 0.43 & 0.46 & 0.72 & 0.650 & 18.34 \\

\textbf{AV-DTEC} \cite{avdtec}& 2024 & Self-Supervised & Audio+Image & 0.33 & 0.25 & 0.27 & 0.58  & \textbf{0.993} & 13.66\\

\textbf{AAUTE} \cite{lei2024audioarraybased3duav}& 2025 & Self-Supervised & Audio-Only & 0.14 & 0.26 & \underline{0.25} & 0.48  & 0.747 & \textbf{59.08}\\

\textbf{YOLOv12} \cite{yolov12} & 2025 & Supervised & Image-Only & 0.21 & 0.52 & 0.33 & 0.54 & 0.620 & 20.11\\

\textbf{YOLO26} \cite{yolo26} & 2026 & Supervised & Image-Only & 0.24 & 0.37 & 0.43 & 0.59 & 0.720 & 24.43\\

\textbf{Lei et al.} \cite{ESWA_Lei} & 2026 & Self-Supervised & Audio+Image & 0.15 & \underline{0.16} & \textbf{0.19} & \textbf{0.29} & 0.560 & 17.80\\

\textbf{Ours}& - & Zero-shot & Image-Only & 0.17 & \textbf{0.15} & 0.44 & \underline{0.30} & \underline{0.960} & 23.50\\
\bottomrule
\end{tabular}
}
\label{tab:3dcomparison}
\\
\vspace{5pt}
\footnotesize{Best results are highlighted in \textbf{bold}, and second best in \underline{underline}. We utilize the zero-shot transfer on the MMAUD to evaluate the accuracy of the trajectory generation process for web videos. This convention is consistently followed throughout the paper. FPS is reported for trajectory estimation modules only.}
\vspace{-2em}
\end{table*}


\subsection{Performance on the MMAUD dataset}


We evaluate our method on the MMAUD dataset against seven representative baselines, as summarized in Table~\ref{tab:3dcomparison}. For YOLO-based approaches, including YOLOv12~\cite{yolov12} and YOLO26~\cite{yolo26}, the corresponding networks are used as feature extractors and augmented with a 3D position regression head to perform trajectory estimation.

To evaluate zero-shot classification performance, each UAV sequence is divided into fixed-length clips of 30 consecutive frames, and clip-level labels are obtained via majority voting. Classification accuracy is computed as the ratio of correctly predicted clips to the total number of clips for each UAV type, as shown in the Table. \ref{tab:3dcomparison}. 

\begin{figure}[t]
\centering
\includegraphics[width=3.1in]{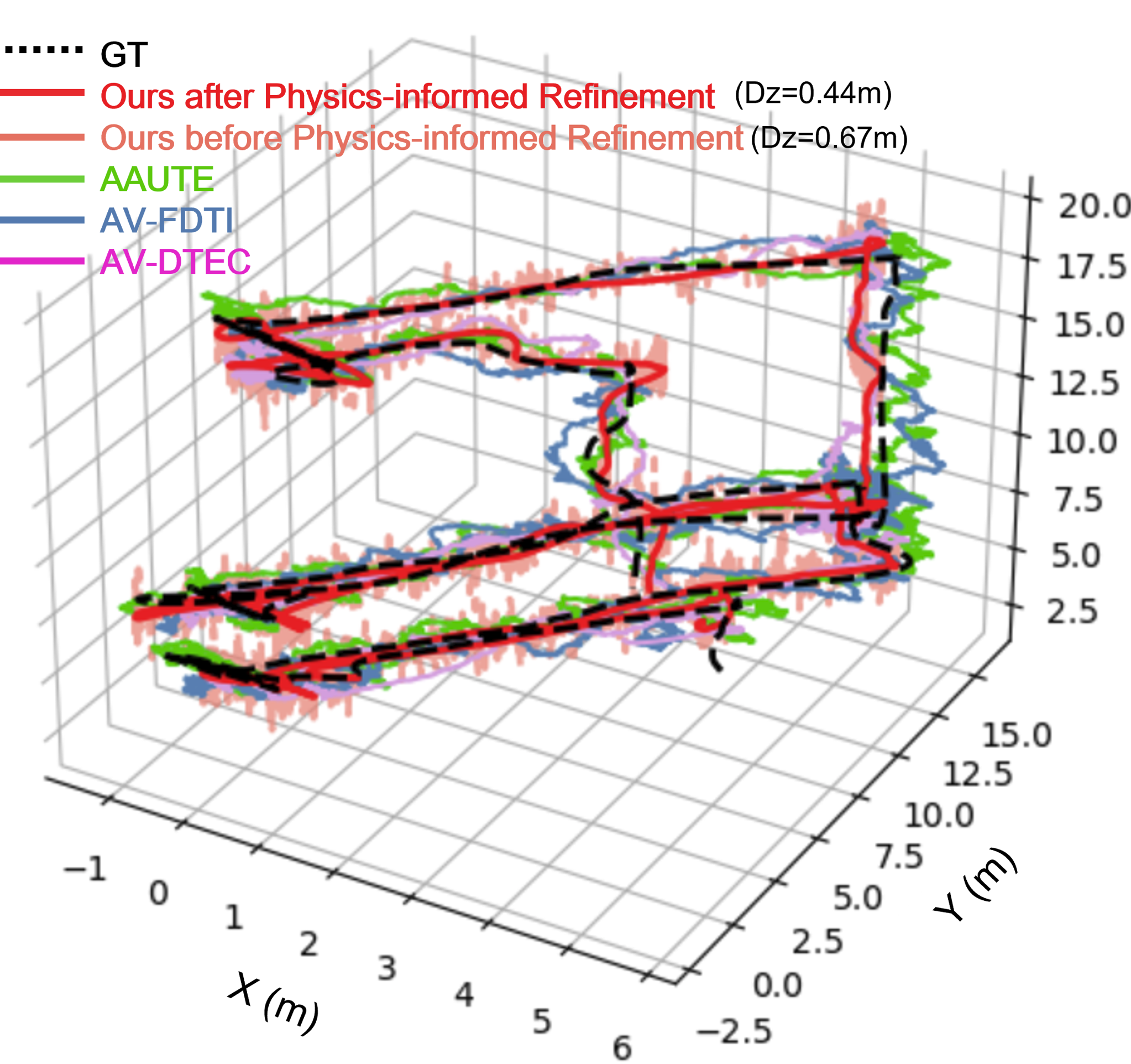}
\caption{Visualization of the comparison results of the DJI Pham4 sequence in MMAUD.}
\label{fig:3d}
\end{figure}



Table~\ref{tab:3dcomparison} reports the quantitative comparison of 3D trajectory estimation. Our method achieves near-SOTA performance under a training-free zero-shot setting, with axis-wise errors ($D_x$, $D_y$, $D_z$) provided for detailed analysis. Fig.~\ref{fig:3d} presents a qualitative comparison on the DJI Pham4 sequence, including ground truth, our results before and after refinement, and open-sourced representative baselines. The proposed method aligns more closely with the ground truth, and EKF refinement notably improves performance along the $Z$ axis (Depth), reducing the error from 0.67 m to 0.44 m.

\subsection{Generalization and Ablation Study}

\begin{table}[h]
\centering
\caption{Generalization of VLM-based video filtering across alternative backbones.
Values denote the number of clips retained from the audited test set after each filtering criterion.}
\label{table:generalization_video_filtering}
\renewcommand{\arraystretch}{1.5}
\setlength{\tabcolsep}{8pt}
\small
\begin{tabular}{ccc}
\hline\hline
\diagbox[width=9em]{Model}{Criterion} & \textbf{Visible UAV} & \textbf{Static Viewpoint} \\ \hline
OpenCLIP (ViT-B/32) & 60 & 21 \\ \hline
SigLIP (ViT-B/32) & 63 & 23\\ \hline
EVA-CLIP (ViT-B/32) & 67 & 27 \\ \hline
MetaCLIP (ViT-B/32) & 64 & 26 \\ \hline
\textbf{CLIP (ViT-B/32)} & \textbf{65} & \textbf{25} \\ \hline
\end{tabular}
\end{table}


To quantitatively examine the generalization of the framework, we construct a controlled set of 200 manually curated video clips, including 50 ``good'' sequences satisfying both visibility and static-view requirements and 150 ``bad'' sequences violating at least one criterion. Manual annotation indicates that 130 clips remain after the visibility filter and 50 after further enforcing the static-view constraint. Applying CLIP to implement the proposed language-driven priors produces retained counts closely aligned with these ground-truth numbers. Replacing CLIP with alternative vision–language models yields comparable results across both filtering stages As summarized in Table~\ref{table:generalization_video_filtering}, different backbones yield comparable retained counts after each filtering stage, indicating that the proposed language-driven priors are not tied to a specific VLM. Similarly, for the classification module in cross-modal label generation, substituting GPT-4o with Qwen2.5-VL-7B and LLaMA-3-8B-Instruct results in classification accuracies of 0.88 and 0.85, respectively. The consistent performance across models further demonstrates that the framework is largely model-agnostic.

Table~\ref{table:ablation_trajectory} presents the ablation study on the trajectory generation component within the cross-modal label generation stage. Using a single expert model leads to relatively large 3D errors (e.g., 0.76 m for Grounding DINO and 0.65 m for Grounded SAM), indicating that individual detectors are insufficient for stable trajectory recovery under zero-shot transfer. Combining multiple experts significantly improves performance. The error decreases to 0.54 m when two complementary experts are fused, demonstrating the effectiveness of cross-expert agreement. With three experts, the error is reduced to 0.30 m, representing a substantial improvement over individual models. Incorporating additional experts ($K=4$ or $K=5$) yields only marginal gains, suggesting diminishing returns as more models are added. Therefore, we adopt $K=3$ as the default configuration, which achieves near-optimal performance while maintaining model simplicity.

\begin{table}[h]
\centering
\caption{Ablation Study on Trajectory Generation in Cross-modal Label Generation Process. We utilize the open-vocabulary model Grounding Dino \cite{dino}, Grounded SAM \cite{groundedsam}, and benchmark solutions from \cite{anti-UAV_dataset} and DUT-AntiUAV \cite{dut-anti}, and a lightweight UAV detector \cite{drone_yolov8}. The evaluation is on the zero-shot transfer on the MMAUD dataset. }
\label{table:ablation_trajectory}
\renewcommand{\arraystretch}{1.5}
\setlength{\tabcolsep}{8pt}
\small
\begin{tabular}{lccccc}
\hline \hline
\cite{dino} & \cite{groundedsam} & \cite{anti-UAV_dataset} & \cite{dut-anti} & \cite{drone_yolov8}  & Error(m)\\ \hline
\checkmark & - & - & - & - & 0.76 \\
- & \checkmark & - & - & - & 0.65 \\
- & \checkmark & \checkmark & - & - & 0.54 \\
- & \checkmark & - & \checkmark & - & 0.59 \\
- & \textbf{\checkmark} & \textbf{\checkmark} & - & \textbf{\checkmark} & \textbf{0.30} \\
- & \checkmark & \checkmark & \checkmark & \checkmark & 0.29 \\
\checkmark & \checkmark & \checkmark & \checkmark & \checkmark & 0.29 \\
\hline

\end{tabular}
\end{table}

\begin{figure}[t]
\centering
\includegraphics[width=3.4in]{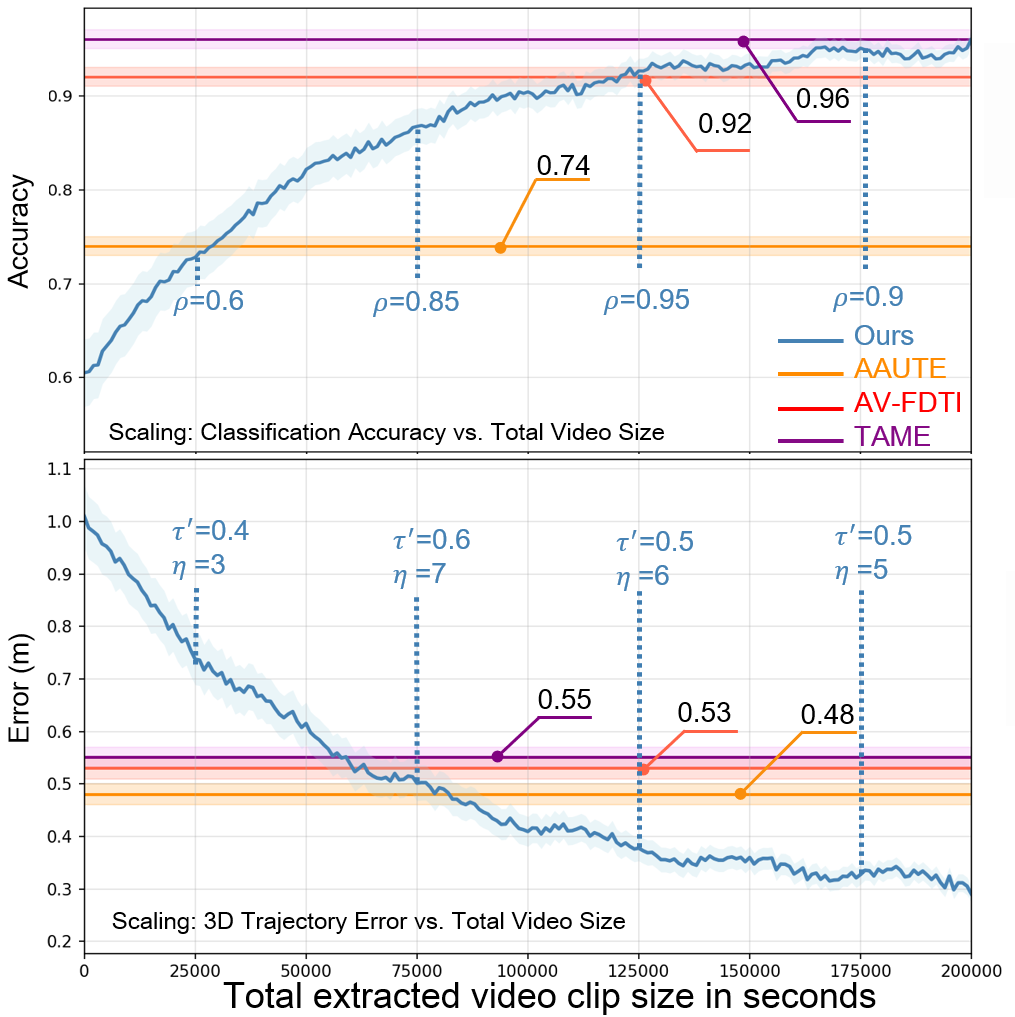}
\caption{Scaling behavior of zero-shot transfer performance.}
\label{scaling}
\end{figure}

\subsection{Data Scaling behavior}


Our framework exhibits a clear data scaling behavior in zero-shot transfer. 
As the amount of web-scale video data increases, we recalibrate several 
hyperparameters in the filtering and refinement stages 
(e.g., majority-voting threshold $\rho$, IoU threshold $\tau'$, B-spline knot spacings $\eta$) based on the accumulated data statistics, the hyperparameters are calibrated solely on the web-scale data. As shown in Fig.~\ref{scaling}, increasing the training corpus to 200{,}000 seconds 
reduces the 3D trajectory error to 0.30\,m and improves classification accuracy to 96\%, 
with zero-shot evaluation on MMAUD, highlighting the scalability and practical potential of the proposed approach.

\begin{figure}[t]
\centering
\includegraphics[width=3.0in]{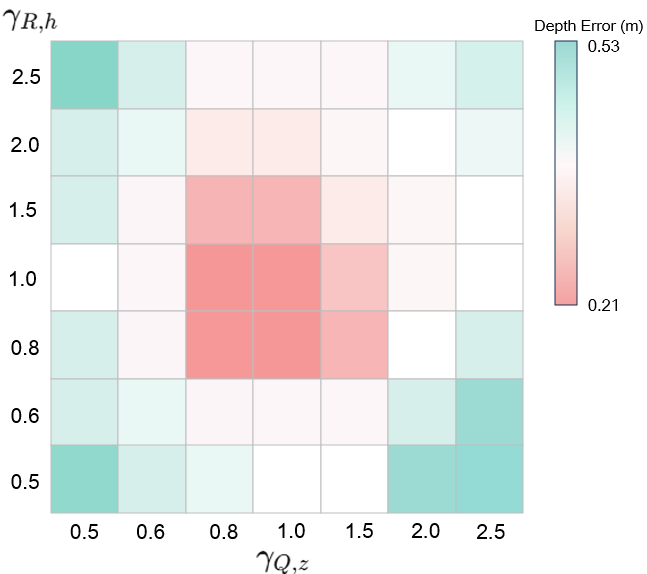}
\caption{ Heatmap for Parameter Sensitivity analysis of Physics-informed refinement process}
\label{fig:sense}
\end{figure}



\subsection{Parameter Sensitivity Analysis}

Parameter sensitivity is critical for achieving stable and robust performance. To examine the influence of noise parameterization in the EKF-based physics-informed refinement module, we introduce two scaling factors to regulate the contribution of depth-related uncertainties. Specifically, the process noise and measurement noise are parameterized as follows:
\begin{equation}
\begin{aligned}
\mathbf Q(\gamma_{Q,z})=\operatorname{diag}\!\Bigl(
    &0.01^2,\;0.01^2,\;\gamma_{Q,z}^2\cdot0.05^2, \\
    &0.10^2,\;0.10^2,\;\gamma_{Q,z}^2\cdot0.50^2
\Bigr),
\end{aligned}
\label{eq:scaledQ}
\end{equation}
\begin{equation}
\mathbf R(\gamma_{R,h})=\operatorname{diag}\!\Bigl(
    0.5^2,\;0.5^2,\;\gamma_{R,h}^2\cdot5.0^2
\Bigr).
\label{eq:scaledR}
\end{equation}
Here, $\gamma_{Q,z}$ scales the process noise of the depth states $(Z,V_z)$, while $\gamma_{R,h}$ scales the measurement noise associated with the bounding-box height $h$. 

Fig.~\ref{fig:sense} illustrates the sensitivity of the scaling factors $\gamma_{Q,z}$ (horizontal axis) and $\gamma_{R,h}$ (vertical axis) via a heatmap, where red regions indicate lower depth errors and blue regions indicate larger errors. The best performance occurs when both parameters are set to $1$. Moreover, depth estimation remains stable within the range $[0.8, 1.5]$, demonstrating the robustness of the EKF-based refinement and highlighting the importance of balancing process and measurement noise in the depth dimension.

\section{Conclusion}

This work introduced a scalable framework for deriving 3D UAV trajectories and type labels from Internet-scale videos without manual annotation. The framework integrated three key components: language-driven acquisition with task-aware filtering, training-free cross-modal label generation combining vision–language reasoning and geometric cues, and physics-informed sequential refinement to ensure temporal consistency and kinematic plausibility. Our results demonstrated favorable data-scaling behavior, showing that freely available web videos can be transformed into reliable image–trajectory pairs for anti-UAV applications. Evaluated via zero-shot transfer on the 3D MMAUD benchmark, the proposed approach achieves competitive performance, closely approaching the current state-of-the-art, highlighting its robustness and practical applicability.

\section*{Acknowledgments} 
The authors acknowledge the use of large language models (e.g., ChatGPT by OpenAI) for assisting in prototype code generation, partial methodology modules, and improving the clarity of the manuscript. All technical contributions and interpretations remain the responsibility of the authors.

\bibliographystyle{IEEEtran}
\bibliography{mybib}

\end{document}